\documentstyle[12pt,pstricks]{article}

\advance\textwidth1in

\def\@citex[#1]#2{\if@filesw\immediate\write\@auxout{\string\citation{#2}}\fi
  \def\@citea{}\@cite{\@for\@citeb:=#2\do
    {\@citea\def\@citea{,\space }\@ifundefined
       {b@\@citeb}{{\bf ?}\@warning
       {Citation `\@citeb' on page \thepage \space undefined}}%
{\csname b@\@citeb\endcsname}}}{#1}}

\newskip\smallskipamount \smallskipamount=3.6pt plus 1.2pt minus 1.2pt
\newskip\medskipamount \medskipamount=7.2pt plus 2.4pt minus 2.4pt
\newskip\bigskipamount \bigskipamount=14.4pt plus 4.8pt minus 4.8pt
\def\smallskip{\vskip\smallskipamount}
\def\medskip{\vskip\medskipamount}
\def\bigskip{\vskip\bigskipamount}

\def\quad{\hskip1em\relax}
\def\qquad{\hskip2em\relax}

\newtheorem{theorem}{Theorem}[section]			%theorems
\newtheorem{lemma}[theorem]{Lemma}			%lemmas
\newtheorem{proposition}[theorem]{Proposition}		%propositions
\newtheorem{definition}[theorem]{Definition}		%definitions
  \newtheorem{eexample}[theorem]{Example}                 %examples
                      {\hfill$\Box$\par\addvspace{\bigskipamount}}

\def\beq{\begin{eqnarray*}}
\def\eeq{\end{eqnarray*}}

 \def\B{{\cal B}} \def\C{{\cal C}} 
   
\def\I{{\cal I}}   
   
   \def\T{{\cal T}}
 \def\V{{\cal V}}

\def\con{\land}

\def\ol#1{\overline{#1}}                %overline
\def\MathCm{5em}

\def\alam#1#2{\lambda{#1}.{#2}}              % lambda abstraction

\def\apc {\mathrel{\wedge}}                  % parallel composition

\def\d={\mathop{:=}}

\def\gleich{\MetaGleich}
\def\MySlash{\hspace{.1ex}\mathpunct/\hspace{-.15ex}}
\def\MyDP{\mathpunct:\hspace{.1ex}}

\def\aex#1#2{\exists{#1}\hspace{.15ex}{#2}}

\def\aexwb#1#2{\aex{#1}{(#2)}}

\def\dnot={\stackrel{.}{\not=}}            % nicht = mit Punkt

\def\gAbstr#1#2#3#4#5{{#1}{#5}{#2}{#4}{#3}}
\def\Abstr#1#2#3{\gAbstr{#1}{#2}{#3}{\MySlash}{\MyDP}}
\def\abstr#1#2#3#4{\Abstr{#1}{{#2}\EMM{#3}}{#4}}
\def\EMM{\hspace{.65mm}}

\def\prod#1{{#1}^{+}}                      % producer
\def\repl#1#2{[#1/#2]}                     % substitutions

\def\rlmodels{\mathrel{\hbox{$\models\kern-.47\MathCm|$}}} % "Aquivalenz
\def\Def#1{{\af{#1}}}

\def\V{{\cal{V}}}

\def\Arrow#1#2#3#4{\mathrel{\stackrel          % -#1->_#2^#3
            {\hspace{-.03\MathCm}{#1}\hspace{.03\MathCm}}{#4}
             \hspace{-.37\MathCm}_{#2}^{#3}}}

\def\orelat#1#2#3{\Arrow{#1}{#2}{#3}           % -#1->_#2^#3
                  {\rightarrow}}
\def\orela#1#2{\orelat{#1}{#2}{}}                % -#1->_#2
\def\rela#1{\orela{}{#1}}                        % ->_#1

\def\rel{\rela{}}                                % ->
\def\congr#1{\mathrel{\equiv_{#1}}}
\def\con{\congr{}}

\def\ol#1{\hspace{.25mm}
          \overline{\hspace{-.25mm}#1\hspace{-.15mm}}
          \hspace{.15mm}}

\def\I{{\cal{I}}}                           % identifiers

\def\Space#1#2{\hspace{#2\MathCm}#1\hspace{#2\MathCm}}  % Space Relation
\def\Sp#1{\Space{#1}{.3}}
\def\hcon{\Space{\con}{.3}}

\def\hmid{\Space{\mid}{.15}}

\newenvironment{Absatz}{\par\vspace{2mm}}
    {\vspace{3.5mm}\mbox{ }\par}

\def\ignore#1{}

\def\Rule#1#2{
    \mbox{$\frac{\;\mbox{$#1$}\;}{\;\mbox{$#2$}\;}$ } }

\def\AxText#1{{({#1})}}
\def\Ax#1{(\mbox{#1})}

\newcommand{\af}[1]{{\sf #1}} % Font f"ur Atome, Pr"afixform
 \def\B{{\cal B}} \def\C{{\cal C}} 
   
\def\I{{\cal I}}   
   
   \def\T{{\cal T}}
 \def\V{{\cal V}}

            {\noindent{\mbox{\bf Proof \mbox{#1}.}}}
           {\hfill$\Box$\par\addvspace{\bigskipamount}}

\def\appl#1#2#3{{#1}\EMM{#2}\EMM{#3}}                  % application

\ignore{}

\def\BV{\B\V}

\def\In{\:\mbox{in}\:}

\def\Arity#1#2{\af{arity}_{#2}(#1)}

\def\Three#1#2#3{{#1}{#2}{#3}}
\def\MS#1#2{\MSA{#1}{}{#2}}
\def\MSA#1#2#3{\Three{#1}{{\subseteq_{{#2}}}} {#3}}

\def\IS#1#2{{#1}\Cap{#2}}

\def\In{\:\mbox{in}\:}
\def\notIn{\:\mbox{not in}\:}

\def\Cap{{\cap}}

\def\gleich{{=}}

\def\Base#1{#1}
\def\Dom{{\sf dom}}

\def\InfTerm{{\IT{V}}}

\newtheorem{expl}{Example}[section]

\def\VarComp#1{\C(#1)}

\def\IT#1{\I\T({#1})}
\def\Def#1{{\em #1}}
\def\Pack#1{\begin{array}{l}{{#1}}\end{array}}
\def\inst#1#2{\mbox{\af{Inst}}_{{#1}}({#2})}
\def\ainst#1#2{\mbox{\af{Inst}'}_{{#1}}({#2})}

\def\BV{\B\V}

           {\hfill$\Box$\par\addvspace{\bigskipamount}}
\def\ISmodels{\mathrel{\models^I}}

\def\ISrlmodels{\mathrel{\rlmodels^I}}

\def\ISSol#1{\mbox{Sol}^I ({#1})}
\def\ISExt#1#2{\mbox{Ext}^I_{{#1}}({#2})}

\newcommand{\DET}{{\circ}}
\newcommand{\Less}[1]{{\le_{{#1}}}}

\def\Feat#1#2{#1[#2]}

\def\AtFeat#1#2#3{#1[#2]#3}
\def\Arity#1#2{#1#2}

\def\WhatsIt{weak subsumption\ }
\newcommand{\weaksub}{\protect{\mbox{$\sqsupset\hspace{-1.07em}\raisebox{-0.5em}{$\sim$}$}}}

\begin{document}

\title{Weak Subsumption Constraints for Type Diagnosis:
       An Incremental Algorithm\\[2mm]
       \large (Extended Abstract)}

\author{Martin M\"uller \hspace{2cm} Joachim Niehren \\
\normalsize        German Research Center for Artificial Intelligence (DFKI)\\
\normalsize        Stuhlsatzenhausweg 3, 66123 Saarbr\"ucken, Germany\\
\normalsize        \{mmueller,niehren\}@dfki.uni-sb.de\\
}
\date{}
\maketitle
\sloppy

\vspace{-1.2cm}

\begin{abstract}
We introduce constraints
necessary for type checking a higher-order concurrent
constraint language,
and solve them with an incremental algorithm.
Our constraint system extends rational unification
by constraints $\MS{x}{y}$
saying that ``$x$ has at least the structure of $y$'',
modelled by a weak instance relation between trees.
This notion of instance has been carefully chosen
to be weaker than the usual one which
renders semi-unification undecidable.
Semi-unification has more than once served to link
unification problems arising from type inference
and those considered in computational linguistics.
Just as polymorphic recursion corresponds to subsumption
through the semi-unification problem, our
type constraint problem corresponds to weak
subsumption of feature graphs in linguistics.
The decidability problem for \WhatsIt for feature
graphs has been settled by D\"orre~\cite{Doerre:WeakSubsumption:94}.
\nocite{RuppRosnerJohnson:94}
In contrast to D\"orre's, our algorithm is fully incremental
and does not refer to finite state automata.
Our algorithm also is a lot more flexible. It allows
a number of extensions (records, sorts, disjunctive types,
type declarations, and others)
which make it suitable for type inference
of a full-fledged programming language.\\[0.2em]

\end{abstract}

\newpage

\section{Introduction}

We give an algorithm which is at the heart of a type
diagnosis system for a higher-order concurrent
constraint language, viz.~the $\gamma$ calculus
\cite{Smolka:GammaCalculus:94} which is the underlying
operational model of the programming language Oz \cite{ozdoc}.
The algorithm decides satisfiability of
constraints
containing equations $x\gleich y$ and $x\gleich f(\ol{y})$,
and \WhatsIt constraints $\MS{x}{y}$ over infinite constructor
trees with free variables.
The algorithm is given fully in terms of constraint simplification.
One the one hand, this gives credit to the close
relationship between type inference and constraint solving
(e.g.,
%% FOLLOWING LINE CANNOT BE BROKEN BEFORE 80 CHAR
\cite{Wand:Simple:87,AikenWimmers:TypeInclusion:93,KozenPalsbergSchwartzbach:JCSS:94}
and many others).
On the other hand it establishes yet another correspondence
between unification problems arising from polymorphic type inference
and unification based grammar formalisms:
The most prominent one is the equivalence of
type checking polymorphic recursion \cite{Mycroft:84,Henglein:88}
with semi-unification \cite{KfouryTiurynUrz:93,DoerreRounds:LICS:90}
 both of which are
undecidable in general. To avoid this undecidability,
we chose a weaker instance relation to give semantics to
$\MS{x}{y}$. For example, we allow $f(a\: b)$ as an instance
of $f(x\: x)$ even if $a\neq b$. On the type
side, this type of constraints maintains some of the polymorphic
flavour, but  abandons full parametric
polymorphism~\cite{MuellerNiehren:Member:94}.

We start out from the set of
infinite constructor trees with holes (free variables).
We give a semantics which interprets the tree assigned
to a variable dually: As itself
and the set of its ``weak'' instances.
Our algorithm terminates, and can be shown to
be correct and complete under this semantics.
The decidability problem for our constraints turned out
to be equivalent to weak subsumption over feature graphs
solved by D\"orre \cite{Doerre:WeakSubsumption:94}
for
feature graphs with feature (but no arity) constraints.

However, only half of D\"orre's two-step
solution is a constraint solving
algorithm. The second step relies on the
equivalence of non-deterministic and deterministic
finite state automata. In contrast, our algorithm
decides satisfiability in a completely incremental
manner and is thus amenable to be integrated
in an concurrent constraint language like Oz~\cite{ozdoc}
or AKL~\cite{JansonHaridi:91}.

The extension of our algorithm towards
feature trees is easily possible
(see \cite{MuellerNiehren:Member:94}).
This allows to do type diagnosis for
records~\cite{SmolkaTreinen:92} and objects.
An entirely set-based semantics allows
to naturally extend the algorithm
to a full-fledged type diagnosis system,
covering -- among other aspects -- sorts, disjunctive types,
and recursive data type declarations
\cite{NiehrenPodelskiTreinen:93}.

{\paragraph{Type Diagnosis.}
As an illustrating example for the form of type diagnosis
we have in mind, consider the following $\gamma$ program:
\[
   \exists x \exists y\exists z\exists p\ \
   \abstr{p}{u}{v}{v\gleich cons(x\: u)} \apc
   \appl{p}{y}{y} \apc x\gleich f(y\: z)
\]
This program declares four variables $x,y,z$, and $p$. It defines
a relational abstraction $p$, which states that its two arguments
$u$ and $v$ are related through the equation $v=cons(x\: u)$.\footnote{
   Note that
   $\abstr{p}{u}{v}{v\gleich cons(x\: u)}$ is different from a
	named $\lambda$ abstraction $p = \alam{u}{cons(x\: u)}$
   because it is relational rather than functional,
   and also different to the Prolog program
   $p(u,v)\ \mbox{:--}\ v=cons(x\: u).$,
   because Prolog does not allow
   variables to be global wrt.~a predicate
   but rather existentially quantifies $x$.
}
Furthermore, it states the equality $x\gleich f(y\: z)$ and
applies $p$ to $yy$.
This application $\appl{p}{y}{y}$ reduces to a copy of
the abstraction $p$ with the
actual arguments $yy$ replaced for the formal ones $uv$:
\[\begin{array}{lcl}
	&   \exists x \exists y\exists z\exists p\ \
   \abstr{p}{u}{v}{v\gleich cons(x\: u)} \apc
   \appl{p}{y}{y} \apc x\gleich f(y\: z) \\
   \to &
   \exists x \exists y\exists z\exists p\ \
   \abstr{p}{u}{v}{v\gleich cons(x\: u)} \apc
   y\gleich cons(x\: y) \apc x\gleich f(y\: z)
\end{array}\]
Observe how the abstraction $p$ is  defined by
reference to the global variable $x$, while
the value of $x$ is defined through an application
of $p$: $\appl{p}{y}{y} \apc x\gleich f(y\: z)$.
Such a cycle is specific to the $\gamma$ calculus
since no other language offers
explicit declaration of logic variables global to an abstraction
(be it logic, functional, or concurrent
languages, e.g., Prolog, ML \cite{HarperMacQueenMilner:86},
or Pict \cite{PierceTurner:Pict:95}).

The types of the variables involved are described by
the following constraint.\footnote{
  The formal account of the derivation of type constraints from programs
  will be given in  \cite{Mueller:96}.
}
For ease of reading, we slightly abuse notation and pick
the type variables identical to the
corresponding object variables:
\[
   p \gleich \langle u\: v\rangle \apc
	v\gleich cons(x\: u) \apc
   \MS{y}{u} \apc \MS{y}{v}  \apc
	x\gleich f(y\:z)
\]
$\langle u\: v\rangle$ is the relational type of $p$,
and the application gives rise to the constraint
$\MS{y}{u} \apc \MS{y}{v}$,
which says that $y$ is constrained by both formal arguments
of the procedure $p$.
The subconstraint
$x\gleich f(y\: z) \apc \MS{y}{v}  \apc v\gleich cons(x\: u)$
reflects the cyclic dependency between $x$ and $p$. It says
that $y$ be in the set of instances of
$v$ which depends through $v\gleich cons(x\: u)$
on $x$, and at the same time that $x$ should be exactly
$f(y\: z)$.

Type diagnosis along this line
is discussed in depth in \cite{MuellerNiehren:Member:94}.

\paragraph{Related Work.}
Apart from the already mentioned work,
related work includes investigations about membership
constraints (e.g., \cite{NiehrenPodelskiTreinen:93}),
type analysis for untyped languages (Soft Typing)
\cite{AikenWimmers:TypeInclusion:93,%
CartwrightFagan:91,WrightCartwright:Scheme:93},
constraint-based program analysis
\cite{KozenPalsbergSchwartzbach:JCSS:94}
and the derivation of recursive sets from programs~\cite{Fruehwirth:91}.
For proofs and a detailed discussion of related work
see~\cite{MuellerNiehren:Member:94}.

\paragraph{Plan of the Paper.}
This paper is structured as follows. In the Section \ref{sec:notation}
below we present our constraints along with their semantics and
give necessary notation. Section \ref{sec:problem} gives a simple
algorithm which is correct but non-terminating.
Section \ref{sec:algorithm} gives the rules of the full algorithm.
Section \ref{sec:outlook} concludes and gives a brief outlook.

\section{Constraints and Semantics}
\label{sec:notation}

We assume a signature $\Sigma$ of function
symbols with at least two elements ranged over by
$f,g,h,a,b,c$ and an infinite
set of {\em base variables\/} $\BV$ ranged over by
$\Base{\chi}$. If $V$ is a further set of variables then
$\InfTerm$ stands for the set of all finite or infinite \Def{trees}
over signature $\Sigma$ and variables $V$. Trees of $\IT{V}$
are always ranged over by $s$ and $t$. The set of variables
occurring in a tree $t$ is denoted by $\V(t)$.
Sequences of variables are written as $\ol{x}$, or $\ol{\chi}$.

We build constraints over a set of
\Def{constraint variables} ranged over by $x$, $y$,
$z$, $u$, $v$, $w$. Constraint variables must contain at least
base variables. The syntax of
our \Def{constraints} $\phi$, $\psi$ is as follows:
\[\begin{array}{rcl@{\qquad \mbox{and} \qquad}rcl}
   x,y & ::= & \chi &
   \phi,\psi  & ::=& x\gleich y \hmid
                     x\gleich f(\ol{y}) \hmid  \MS{x}{y} \hmid
	              \phi \apc \psi
\end{array}\]
As {\em atomic constraints\/} we consider equations
$x\gleich y$ or $x\gleich f(\ol{y})$ and \WhatsIt
constraints $\MS{x}{y}$. Constraints are atomic
constraints closed under conjunction. \Def{First-order
formulae} build over constraints $\phi$ are denoted by
$\Phi$. We define $\con$ to be the least binary
relation on $\phi$ such that $\apc$ is associative
and commutative. For convenience, we shall use the
following notation:
\[
\begin{array}{r@{\qquad}c@{\quad}l}
  \mbox{$\phi$ in $\psi$}& \mbox{iff} &\mbox{exists $\phi'$ with
                          $\phi\apc\phi' \con \psi$}
\end{array}
\]
As semantic structures we pick \Def{tree-structures}
which we also call $\IT{V}$ for some set $V$. The domain of a
tree-structure $\IT{V}$ is the set of trees $\IT{V}$.
Its interpretation is defined by
$\: f^{\IT{V}}(\ol{t}) = f(\ol{t})$. We define the application
$f(\ol{T})$ of $f$ to a sequences of sets of trees $\ol{T}$
elementwise, $f(\ol{T})= \{f(\ol{t}) \mid \ol{t}\in \ol{T}\}$.
Given a tree $s\in \IT{V}$, the set $\inst{V}{s}$
of \Def{weak instances of $s$} is defined as the greatest fixed point of:
\[
\inst{V}{s} = \left\{\begin{array}{ll}
     \IT{V} & \mbox{if $t = x$ for some $x$} \\
     f(\ol{\inst{V}{s}}) &  \mbox{if $t = f(\ol{s})$ for some $\ol{s}$}
    \end{array}\right.
\]
Notice that this definition implies
$f(a\ b) \in \inst{V}{f(x\ x)}$, even if $a\not= b$.
Let $V_1$, $V_2$ be two sets whose elements we call
variables. A \Def{$V_1$-$V_2$-substitution} $\sigma$ is a
mapping from $V_1$ to $\IT{V_2}$. By homomorphic extension, every
substitution can be extended to a mapping from $\IT{V_1}$
to $\IT{V_2}$.
The set of \Def{strong instances of $s$}
is defined by
$
\ainst{V}{s} =
  \{\sigma(s) \:|\: \mbox{$\sigma$ is a $\V(s)$-$V$-substitution}\}
$.
Note that $\ainst{V}{s} \subseteq \inst{V}{s}$, and that
$f(a\: b)\not\in \ainst{V}{f(x\: x)}$ if $a\neq b$.
Using $\ainst{V}{s}$ instead of $\inst{V}{s}$ would
make satisfiability of our constraints equivalent to
semi-unification and undecidable
\cite{Kfoury:SemiUnification:90,DoerreRounds:LICS:90}.

Let $\sigma$ be a $V_1$-$V_2$-substitution,
$\{x,y,\ol{z}\}\subset V_1$, and $\phi,\psi$
constraints such that $\V(\phi)\subseteq V_1$, $\V(\psi)\subseteq V_1$.
Then we define:\label{semantics}
\[\begin{array}{lcl@{\qquad\qquad} lcl}
  \models_{\sigma} x\gleich  y &
         \mbox{iff} & \sigma(x) \gleich  \sigma(y) &
  \models_{\sigma} x\gleich  f(\ol{z})&
          \mbox{iff} & \sigma(x) \gleich  f^{\IT{V_2}} (\ol{\sigma(y)}) \\
  \models_{\sigma} \MS{x}{y} &
          \mbox{iff} & \inst{V_2}{\sigma(x)}\subseteq \inst{V_2}{\sigma(y)} &
  \models_{\sigma} \phi \wedge \psi & \mbox{iff} &
  \models_{\sigma} \phi \mbox{ and }  \models_{\sigma} \psi
\end{array}\]

A \Def{$V_1$-$V_2$-solution} of $\phi$ is a $V_1$-$V_2$-substitution
satisfying $\models_{\sigma} \phi$. A constraint $\phi$ is called
\Def{satisfiable}, if there exists a $V_1$-$V_2$-solution for
$\phi$. The notion of $\models_\sigma$ extends to arbitrary first-order
formulae $\Phi$ in the usual way.
We say that a formula $\Phi$ is \Def{valid}, if $\models_\sigma
\Phi$ holds for all $V_1$-$V_2$-substitutions $\sigma$ with
$\V(\Phi)\subseteq V_1$. In symbols, $\models \Phi$.

Our setting is a conservative extension of
the usual rational unification problem. This means
that free variables in the semantic domain do not affect
equality constraints. A constraint $\phi$ is \Def{satisfiable in the
tree-model $\IT{V}$}, if there
exists a $\BV$-$V$-solution of $\phi$. The trees of
$\IT{\emptyset}$ are called \Def{ground trees}.

\begin{proposition}
\label{as usual}
Suppose $\phi$ not to contain \WhatsIt constraints.
Then $\phi$ is satisfiable if and only if it is satisfiable in
the model of ground trees.
\end{proposition}

The statement would be wrong for $\phi$'s
containing weak subsumption constraints. For instance,
consider the following $\phi$ with $a\not= b$:
\[
  \phi \hcon \MS{x}{z} \apc \MS{y}{z} \apc x\gleich a \apc y\gleich b\:
\]
This $\phi$ is not satisfiable in the model of ground trees,
since the set $\inst{\emptyset}{t}$ is a singleton for
all ground trees $t$, whereas any $V_1$-$V_2$-solution $\sigma$ of
$\phi$ has to satisfy
$
   \{a,b\} \subseteq \inst{V_2}{\sigma(z)}
$.
However, there exists a $\{x,y,z\}$-$\{v\}$-solution
$\sigma$ of $\phi$, where $\{v\}$ is an singleton:
$
  \sigma(x) \Sp{=} a\:,
  \sigma(y) \Sp{=} b\:,
  \sigma(z) \Sp{=} v\:
$.

\begin{proposition}\label{lemma-eq-implies-in}
For all $x$, $y$, $z$, $u$, $v$ the following statements hold:
\[ \begin{array}{ll@{\qquad}ll@{\qquad}ll}
   1) &\models\: x\gleich y \rel \MS{x}{y}\:, &
   2) &\models\: \MS{x}{y}\apc \MS{y}{z} \rel \MS{x}{z}\:, &
   3) &\models\: x\gleich f(\ol{y}) \rel \MS{x}{f(\ol{y})}\:, \\
   4) &\not\models\: \MS{x}{y} \apc \MS{y}{x} \rel x\gleich y &
   5) &
  \multicolumn{3}{l}{
    \not\models\:  x\gleich f(u\ v) \apc \MS{x}{y} \apc y\gleich f(z\ z)
	       \rel u\gleich v\:.
  }
  \end{array}\]
\end{proposition}

\paragraph{Weak Subsumption vs.~Sets of Weak Instances.}
In the remainder of this section we
compare our sets of weak instance with D\"orre's
notion of weak subsumption.
Let us consider constructor
trees as special feature trees with integer-valued features,
a distinguished feature
{\sf label}~(e.g., \cite{NiehrenPodelski:93,Backofen:94}),
and a distinguished feature {\sf arity}.
Given feature constraints $x[f]y$ saying that $x$
has direct subtree $y$ at feature $f$, the equation
$x\gleich f(y_1\dots y_n)$ can be considered
equivalent to:\footnote{
  This simpler encoding of constructor trees not using arity
  constraints has been suggested by one of the referees.
}
\[
       x[{\sf arity}]n
       \apc x[{\sf label}]f \apc x[1]y_1 \apc \dots \apc x[n]y_n.
\]
Let us write $s[f]\!\!\downarrow$ to say that
the tree $s$ has some direct subtree at $f$.
A \Def{simulation} between $\IT{V_1}$ and $\IT{V_2}$ is a relation
$\Delta \subseteq \IT{{V_1}} \times \IT{{V_2}}$ satisfying:
If $(t,s) \in \Delta$ then
\begin{center}
\begin{tabular}{lp{10cm}}
\Ax{Arity Simulation} & If $t[{\sf label}]\!\!\downarrow$
	and there is an $n$ such that $t[{\sf arity}]n$,
	then $s[{\sf arity}]n$.\\
\Ax{Feature Simulation} &
   If $t[f]\!\!\downarrow$ and there is a tree $t'$
	such that $t[f]t'$,
	then
   $s[f]\!\!\downarrow$, $s[f]s'$, and
	$(t',s') \in \Delta$.
\end{tabular}
\end{center}

Now, the weak subsumption preorder $\weaksub^V$ is defined by:
\[
\mbox{$t \weaksub^V s$\qquad iff\qquad
 there is a simulation  $\Delta\subseteq V\times V$ such that
$(s,t)\in\Delta$}
\]
We have the following lemma:
\begin{lemma}
For all constructor trees $s,t$ it holds that:
$\inst{V}{s} \subseteq \inst{V}{t}$ iff $s \weaksub^V t$.
\end{lemma}
A similar statement can be derived for the set of strong
instances
and a strong subsumption preorder
following~\cite{Doerre:WeakSubsumption:94}.
The difference between $\weaksub^V$ and D\"orre's
notion of weak subsumption is that he does not require
\AxText{Arity Simulation}, while we naturally do
since we  start
from constructor trees. For type checking, constructor
trees seem more natural: For illustration note that
the arity of a procedure is essential type information.

\section{A Non-terminating Solution}
\label{sec:problem}

\begin{figure}[htb]
\framebox[\textwidth]{
$\begin{array}{lll}
\Ax{Decom}&
\Rule{x\gleich f(\ol{u}) \apc \phi}
     {\ol{u}\gleich \ol{v} \apc \phi}
& \Pack{x\gleich f(\ol{v}) \In \phi.}
\vspace{4mm}\\
\Ax{Clash}&
\Rule{x\gleich f(\ol{u})  \apc \phi}
     {\bot}
& \Pack{x\gleich g(\ol{v}) \In \phi,\: and\:  f\not= g.}
\vspace{4mm}\\
\Ax{Elim}&
\Rule{x\gleich y \apc \phi}
     {x\gleich y \apc \phi\repl{y}{x}}
& \Pack{ x\in\V(\phi),\mbox{ and } x\not= y.}
\vspace{4mm}\\
\Ax{Descend} &
\Rule{\MS{x}{y}\apc \phi} {x\gleich f(\ol{u}) \apc
      \MS{\ol{u}}{\ol{z}}\apc \phi}
& \begin{array}{l}
  \ol{u} \mbox{ fresh}, y\gleich f(\ol{z}) \In \phi.
  \end{array}
\end{array}$
}
\caption{\label{fig:loop-algorithm}A Non-Terminating Algorithm}
\end{figure}

In order to solve our constraints one could come up
with the system
given in Figure \ref{fig:loop-algorithm}.
Besides the three usual unification rules for
rational trees, the only additional rule is (Descend).
This algorithm is correct and very likely to be complete
in that for an unsatisfiable constraint $\phi$ there is a derivation
from $\phi$ to $\bot$.
However, this intuitive algorithm loops due to the introduction
of new variables.
\[
 \begin{array}{c@{\qquad}l@{\qquad}c}
    \underline{\MS{x}{y} \apc y \gleich f(x)}
      & \raisebox{-0.5em}{Descend} &
    \underline{\MS{x}{y} \apc y \gleich f(y)}
 \\
    \underline{ x\gleich f(x_1)\apc \MS{x_1}{x}  \apc y \gleich f(x)}
      & \raisebox{-0.5em}{Descend} &
    \underline{ x\gleich f(x_1)\apc \MS{x_1}{y} \apc y \gleich f(y)}
  \\
       \dots &&\ldots
  \end{array}
\]
Note that some form of descending is necessary in order to derive the clash
from the inconsistent constraint
\mbox{$
  y \gleich f(u) \apc u\gleich a \apc z \gleich f(x) \apc
  \MS{x}{y} \apc \MS{x}{z} \apc \phi
$}

\section{Algorithm}
\label{sec:algorithm}

To consider trees with free variables as set of instances
means that we need to compute intersections of such sets
and to decide their emptiness.
When we simplify $\MS{x}{y}\apc \MS{x}{z}$
in a context $\phi$, we have to compute the intersection of the
sets of instances of $y$ and $z$.
In order to avoid the introduction of new variables we
add a new class of variables to represent such intersections,
and one new constraint.
\Def{Intersection variables} are defined as nonempty
finite subsets of base variables. In order
capture the intended semantics, we
write $\chi_1\Cap\ldots\Cap \chi_n$ instead of
$\{\chi_1\}\cup \ldots \cup\{\chi_n\}$.
The equality $\con$ on intersection variables is the equality
on powersets, which  satisfies:
\[
     x\Cap y \con y \Cap x, \quad
     (x\Cap y)\Cap z \con x\Cap (y\Cap z),\quad
    x\Cap x \con x.
\]
We call an $x$ a \Def{component} of $y$, if
$y\equiv \IS{x}{z}$ for some $z$.
The set of components of a variable $x$ is denoted by
$\VarComp{x}$. Note that \mbox{$\IS{x}{y}\in \V(\phi)$} implies
$x\in\VarComp{\V(\phi)}$ but in general not $x\in\V(\phi)$.
As additional constraint we introduce $\MS{x}{f(\ol{y})}$,
with the semantics:
\[
\models \: \MS{x}{f(\ol{y})} \leftrightarrow
        \aexwb{u}{\MS{x}{u}\apc u\gleich f(\ol{y})}\:.
\]
Complete semantics has to take care of
intersection variables such as $y\Cap z$.
Constraint solving will propagate intersection variables
into most constraint positions. That is, our algorithm
actually operates on the following constraints:
\[\begin{array}{rcl@{\qquad \mbox{and} \qquad}rcl}
   x,y & ::= & \chi \hmid x\Cap y &
   \phi,\psi  & ::=& x\gleich y \hmid
                     x\gleich f(\ol{y}) \hmid  \MS{x}{y} \hmid
		     \MS{x}{f(\ol{y})} \hmid
	              \phi \apc \psi
\end{array}
\]
However, if started with a constraint containing
only base variables, our algorithm maintains
this invariant for the equational constraints.

Let us call a variable $x$ \Def{immediately determined by f},
in $\phi$, written $x \DET f(\ol{y})$, if one of
$   x \gleich f(\ol{y}) $ or $\MS{x}{f(\ol{y})} $
is in $\phi$ for some $f(\ol{y})$.
We say that $x$ is \Def{immediately determined} in $\phi$
if it is immediately determined by some $f$ in $\phi$.
Call $x$ \Def{determined}, written
$x \Less{\phi} f(\ol{u})$ if $x$ is immediately determined
in $\phi$, or
$  \MS{x}{\IS{y}{z}}$ and $y \DET f(\ol{u})$ are in $\phi$.
Obviously, if $x\Less{\phi} f(\ol{y})$, then the top-level
constructor of $x$ must be $f$.

We define the application of an operator
$\repl{y}{x}$ to intersection variables,
only if $x$ is a base variable. If
$z \con (\chi_1\Cap \ldots \Cap \chi_n)$, then
$z\repl{y}{x}$ we define:
\[
    z\repl{y}{x} \con \chi_1\repl{y}{x}\Cap\ldots\Cap\chi_n\repl{y}{x}\:.
\]
We say that $\repl{y}{x}$ applied to
intersection variables performs \Def{deep substitution}.
The following property holds for deep substitution:
\[
   \C(\V(x\gleich y \apc \phi))  \Sp{=} \C(\V(x\gleich y \apc \phi[y/x]))\:.
\]
Note however that $\:\V(x\gleich y \apc \phi) \not=
     \V(x\gleich y \apc \phi[y/x])$ if $\phi \con \MS{z}{x\Cap y}$.
The variable $\IS{x}{y}$ is contained in the first but not in
the second set.
We can now specify our algorithm for constraint simplification.
It is given by the rules in Figure \ref{fig:unification} and
Figure \ref{fig:algorithm}.

\begin{figure}[hbt]
\framebox[\textwidth]{
$\begin{array}{lll}
\Ax{Decom}&
\Rule{x\gleich f(\ol{u}) \apc \phi}
     {\ol{u}\gleich \ol{v} \apc \phi}
& \mbox{$x\gleich f(\ol{v})$ in $\phi$.}
\vspace{4mm}\\
\Ax{Clash}&
\Rule{\phi}
     {\bot}
&   x\Less{\phi} f(\ol{u}),\: x\Cap y \Less{\phi} g(\ol{v}),\:
        \mbox{and }f\not= g.
\vspace{4mm}\\
\Ax{Elim}&
\Rule{x\gleich y \apc \phi}
     {x\gleich y \apc \phi\repl{y}{x}}
&  x\in\VarComp{\V(\phi)}\cap\BV,\mbox{ and } x\not\equiv y.
\end{array}$
}
\caption{\label{fig:unification}Rational Tree Unification}
\end{figure}

The Rule \AxText{Decom} is known from usual unification
for rational trees. Up to the application condition
$x\in\VarComp{\V(\phi)}\cap\BV$, this also applies to rule
\AxText{Elim}. This side condition accounts for
deep substitution. The \AxText{Clash} rule contains
as special cases:
\[
   \Rule{x\gleich f(\ol{y}) \apc x\gleich g(\ol{z}) \apc \phi}{\bot}
 f\not= g
\qquad\mbox{and}\qquad
 \Rule{\MS{x}{f(\ol{y})} \apc \MS{x}{g(\ol{z})} \apc \phi}{\bot}
 f\not= g\:.
\]
Its full power comes in interaction with the rules in
Figure \ref{fig:algorithm}. Then it allows to derive a clash
if for a variable $x$ a constructor is known, and for
some variable $\IS{x}{y}$ a distinct constructor
is derivable.

\begin{figure}[htb]
\framebox[\textwidth]{
$\begin{array}{lll}
\Ax{Propagate1}&
\Rule{\MS{\IS{x}{y}}{z} \apc \phi}
     {\MS{\IS{x}{y}}{z\Cap u} \apc\phi}
& \begin{array}{l}
   \MS{x}{u}\In\phi,\:z\Cap u\not\equiv z.
  \end{array}
\vspace{4mm}\\
\Ax{Propagate2}&
\Rule{\MS{\IS{x}{y}}{f(\ol{u})} \apc \phi}
     {\MS{\IS{x}{y}}{f(\ol{u}\Cap\ol{v})} \apc \phi}
& \begin{array}{l}
   x\Less{\phi}f(\ol{v}),\:\ol{u}\Cap\ol{v}\not\equiv \ol{u}.
  \end{array}
\vspace{4mm}\\
\Ax{Collapse}&
\Rule{\MS{x}{\IS{y}{u}} \apc \phi}
     {\MS{x}{\IS{y}{\IS{z}{u}}} \apc \phi}
&\begin{array}{l}
  \MS{y}{z} \In \phi,
  \mbox{ and $\IS{y}{\IS{z}{u}} \not\equiv \IS{y}{z}$.}
 \end{array}
\ignore{
\vspace{4mm}\\
\Ax{Intersect1}&
\Rule{\MS{x}{y}\apc \MS{x}{z}\apc \phi}
     {\MS{x}{\IS{y}{z}} \apc\phi}
\vspace{4mm}\\
\Ax{Intersect2}&
\Rule{\MS{x}{f(\ol{y})}\apc \MS{x}{f(\ol{z})}\apc \phi}
     {\MS{x}{f(\IS{\ol{y}}{\ol{z}})} \apc\phi}
}%ignore
\vspace{4mm}\\
\Ax{Descend1}&
\Rule{x \gleich f(\ol{u}) \apc \phi}{x \gleich f(\ol{u}) \apc
\MS{\ol{u}}{\ol{v}} \apc \phi}
&    \begin{array}{l}
     x \Less{\phi} f(\ol{v}), \\
     \MS{\ol{u}}{\IS{\ol{v}}{\ol{w}}} \mbox{ not in } \phi
     \end{array}
\vspace{4mm}\\
\Ax{Descend2}&
\Rule{\phi}
     {\MS{\IS{x}{y}}{f(\ol{u})} \apc \phi}
&\begin{array}{l}
          \IS{x}{y}\in\V(\phi),\: x\Less{\phi} f(\ol{u}),\\
          \mbox{and not }
%	\IS{x}{y}\Less{\phi} {g(\ol{v})}.
	\IS{x}{y}\DET g(\ol{v}) \In \phi.
     \end{array}
\end{array}$
}
\caption{\label{fig:algorithm}Simplifying Membership Constraints}
\end{figure}

Rules \AxText{Propagate1} and \AxText{Propagate2}
propagate intersection variables into the right
hand side of \WhatsIt contraints. The \AxText{Collapse}
rule collapses chains of variables related via
\WhatsIt constraints. In other words, these rules propagate
lower bounds with respect to the \WhatsIt relation.

The rules \AxText{Descend1} and \AxText{Descend2} replace
\AxText{Descend} from the non-terminating algorithm
in Figure \ref{fig:loop-algorithm}. The Descend rules
are the only rules  introducing new \WhatsIt constraints.
The rule \AxText{Descend2} introduces a constructor
for intersection-variables $x\Cap y$ by adding a
constraint of the form $\MS{x\Cap y}{f(\ol{u})}$.
If the rule is applied, then the intersection
of $x$ and $y$ is forced to be nonempty. Nonemptiness
is implied by $\phi$, if $\IS{x}{y}$ occurs in $\phi$
($\IS{x}{y}\in\V(\phi)$).

Note that \AxText{Descend1} and \AxText{Descend2} are
carefully equipped with side conditions for termination.
For example, the following derivations are {\bf not}
possible:
\[
   \Rule{x\gleich f(u)}
        {x\gleich f(u) \apc \MS{x}{f(u)}}
\qquad
   \Rule{\MS{x}{y} \apc x\gleich f(x) \apc \MS{x}{f(y)}}
        {\MS{x}{y} \apc \MS{x}{y} \apc x\gleich f(x) \apc \MS{x}{f(y)}}
\qquad
\Rule{x\gleich f(y)}
     {\MS{y}{y} \apc x\gleich f(y)}\:.
\]

We can prove that our algorithm
performs equivalence transformations with respect to
substitutions $\sigma$ which meet the intended
semantics of intersection variables, i.e.,
\Def{intersection-correct} substitutions:

\begin{definition}[Intersection Correct]
We say that a substitution $\sigma$ is
\Def{intersection-correct for $x$ and $y$},
if it satisfies:
\[
  \sigma(\IS{x}{y}) =  \sigma(x) \cap \sigma(y)\:.
\]
We say that a substitution $\sigma$ is \Def{intersection-correct},
if the following properties holds for all intersection
variables $x,y$ and $z$:
\[
 \begin{array}{l}
  \mbox{If $x$, $y$, $\IS{x}{y}$ $\in$ $\Dom(\sigma)$,
	then $\sigma$ is intersection-correct for $x$ and $y$}.\\
  \mbox{If $x$, $\IS{x}{y}$ $\in$ $\Dom(\sigma)$, then $\sigma$
        is intersection-correct for $\IS{x}{y}$ and $y$}.
  \end{array}
\]
\end{definition}
Note that $\sigma$ is intersection-correct for $x$ and $x\Cap y$, iff
$\sigma(\IS{x}{y}) \Sp{\subseteq} \sigma(x)$.
We call a constraint $\phi$
\Def{intersection-satisfiable}, if $\phi$ has
an intersection-correct solution.

\begin{proposition}
\label{satintsat}
Let $\phi$ be a constraint only containing base variables only.
Then $\phi$ is satisfiable, if and
only if it is intersection satisfiable.
\end{proposition}

We denote the set of all intersection-correct
solutions of $\phi$ with $\ISSol{\phi}$.
Assume $\sigma$ to be a substitution. A
\Def{$V$-extension} of $\sigma$
is a substitution $\tilde{\sigma}$ such that
$\Dom{(\tilde{\sigma})} = \Dom{(\sigma)} \cup V$ such
that $\sigma$ and $\tilde{\sigma}$ coincide on $\Dom{(\sigma)}$.
We denote the set of all intersection-correct $V$-extensions
of $\sigma$ with $\ISExt{V}{\sigma}$.
Let $\phi$ and $\psi$ be constraints. We say that
$\phi$ \Def{intersection-implies} $\psi$, written
$\phi\ISmodels \psi$, if
\[
 \ISExt{\V(\psi)} {\ISSol{\phi}} \Sp{\subseteq} \ISSol{\psi}
 \ \ \ \mbox{ and }\ \ \
	\ISSol{\phi}=\emptyset \mbox{ iff }
	\ISExt{\V(\psi)}{\ISSol{\phi}}=\emptyset
\]
We call $\phi$ and $\psi$ \Def{intersection-equivalent} if
$\phi \ISmodels\psi$ and $\psi \ISmodels\phi$, and
 write $\phi \ISrlmodels \psi$.
Both conditions ensure the following Lemma:

\begin{lemma}
If $\phi$ is  not intersection
satisfiable, then $\phi\ISmodels\psi$ holds vacuously for all $\psi$.
Furthermore, if $\phi \ISrlmodels \psi$, then
$\phi$ is intersection satisfiable if and only if
$\psi$ is.
\end{lemma}

Given the above notions, the following two theorems
are our main results. For the proofs the reader is referred
to \cite{MuellerNiehren:Member:94}.

\begin{theorem}[Termination]
\label{Termination}
The rule system given in Figures
\ref{fig:unification} and \ref{fig:algorithm}
terminates.
\end{theorem}

\begin{theorem}[Correctness and Completeness]
\label{CorCom}
Let $\phi$ be a constraint containing base variables only.
Then the following statements are equivalent:
\begin{enumerate}
\item
$\phi$ is intersection-satisfiable.
\item
There exists an irreducible $\psi\neq \bot$ derivable
from $\phi$.
\item
There exists a irreducible $\psi\neq \bot$ that is intersection-equivalent
to $\phi$.
\item
$\bot$ cannot be derived from $\phi$.
\end{enumerate}
\end{theorem}

\paragraph{Acknowledgements.}

We would like to thank Ralf Treinen for pointing
us to D\"orre's paper and the anonymous referees
for useful remarks.
The research reported in this paper has been supported by the
Bun\-des\-mi\-ni\-ster f\"ur Bildung, Wissenschaft, Forschung und
Technologie (FTZ-ITW-9105), the Esprit Project \hbox{ACCLAIM} (PE~7195),
the Esprit Working Group CCL (EP~6028), and a fellowship of the
Graduiertenkolleg 'Kognition' at the Universit\"at des Saarlandes
of the first author.
\section{Outlook}
\label{sec:outlook}

\ignore{
\begin{figure}[htb]
\framebox[\textwidth]{
$\begin{array}{lll}
\Ax{FeatProp} &
\Rule{ \phi }{ \phi \apc \Feat{x}{f} } &
   \begin{array}{l}
     \MS{x}{y}\In \phi,
	\Feat{y}{f} \In \phi \\
        \Feat{x}{f} \notIn \phi
   \end{array}
\vspace{4mm}\\
\Ax{FeatDecom} &
\Rule{\AtFeat{x}{f}{y} }{ y\gleich z } &
  \begin{array}{l}
    \AtFeat{x}{f}{z} \In \phi
  \end{array}
\vspace{4mm}\\
\Ax{ArityProp} &
\Rule{ \phi }{ 	\phi \apc \Arity{x}{F} } &
   \begin{array}{l}
     \MS{x}{y},
	\Arity{y}{F} \In \phi \\
        \Arity{x}{F} \notIn \phi
   \end{array}
\vspace{4mm}\\
\Ax{ArityDef} &
\Rule{ \phi }{ 	\phi \apc \Arity{x}{F} } &
   \begin{array}{l}
        x\DET f(a_1:x_1 \dots a_n:x_n), F=\{a_1,\dots,a_n\} \\
	\Arity{x}{F} \notIn \phi
   \end{array}
\end{array}$
}
\caption{\label{fig:records}Handling Records}
\end{figure}
}

We have presented an algorithm for deciding
satisfiability of \WhatsIt constraints over infinite
constructor trees with holes.
Our motivation to solve such constraints
grew out of a type inference problem. Formally,
the problem is equivalent to type checking a weak form
of polymorphic recursion. Type checking
polymorphic recursion is equivalent to
semi-unification and to subsumption of feature graphs.
All three are undecidable
\cite{Henglein:88,KfouryTiurynUrz:93,DoerreRounds:LICS:90}.
We establish a similar correspondence between
a type inference problem and
weak subsumption of feature graphs:
The latter has been investigated by D\"orre
looking for a logical treatment
of coordination phenomena in unification based
grammar formalisms \cite{Doerre:WeakSubsumption:94}.
Our starting point from the constraint language Oz however
lead us to an incremental algorithm, in contrast
to the automata based solution of D\"orre.

\end{document}